\documentclass[10pt,twocolumn,letterpaper]{article}

\usepackage{iccv}
\usepackage{times}
\usepackage{epsfig}
\usepackage{graphicx}
\usepackage{amsmath}
\usepackage{amssymb}
\usepackage{amsmath,amssymb} % define this before the line numbering.
\usepackage{pifont}
\usepackage{amsfonts} 
\usepackage{graphicx}
\usepackage{soul}
\usepackage{url}
\usepackage[utf8]{inputenc}
\usepackage{booktabs}
\usepackage{algorithm}
\usepackage{algorithmic}
\usepackage[shortlabels,inline]{enumitem}
\usepackage{comment}
\usepackage{color}
\usepackage{xcolor}
\usepackage{subfigure}
\usepackage{array}
\newcolumntype{H}{>{\setbox0=\hbox\bgroup}c<{\egroup}@{}}
\usepackage{enumitem}
\newcommand{\cmark}{\ding{51}}%
\newcommand{\xmark}{\ding{55}}

\usepackage{amsthm}
\theoremstyle{definition}
\newtheorem{theorem}{Theorem}[section]
\newtheorem{definition}{Definition}[section]
\newtheorem{lemma}[theorem]{Lemma}

\usepackage[hidelinks]{hyperref}

\iccvfinalcopy % *** Uncomment this line for the final submission

\def\iccvPaperID{2} % *** Enter the ICCV Paper ID here

\ificcvfinal\fi

\begin{document}

%%%%%%%%% TITLE
\title{SCARLET-NAS: Bridging the Gap between Stability and Scalability in Weight-sharing Neural Architecture Search}

\author{
Xiangxiang Chu\thanks{This work was done when all the authors were at Xiaomi AI Lab.} \quad Bo Zhang \quad Qingyuan Li \quad Ruijun Xu \quad Xudong Li \\
{\small \tt \{chuxiangxiang,zhangbo11,liqingyuan, xuruijun\}@xiaomi.com, lixudong16@mails.ucas.edu.cn}
}
%% omit the following lines up until the closing ``}''.
%% Additional authors and addresses can be added with ``\and'',
%% just like the second author.
%% To save space, use either the email address or home page, not both
%\and

\maketitle
\ificcvfinal\thispagestyle{empty}\fi

%%%%%%%%% ABSTRACT
\begin{abstract}
To discover powerful yet compact models is an important goal of neural architecture search. Previous two-stage one-shot approaches are limited by search space with a fixed depth. It seems handy to include an additional skip connection in the search space to make depths variable. However, it creates a large range of perturbation during supernet training and it has difficulty giving a confident ranking for subnetworks. In this paper, we discover that skip connections bring about significant feature inconsistency compared with other operations, which potentially degrades the supernet performance. Based on this observation, we tackle the problem by imposing an \emph{equivariant learnable stabilizer} to homogenize such disparities (see Fig.\ref{fig:spos-fairnas-cosine-angle-3rd-els}).  Experiments show that our proposed stabilizer helps to improve the supernet's convergence as well as ranking performance. With an evolutionary search backend that incorporates the stabilized supernet as an evaluator, we derive a family of state-of-the-art architectures, the SCARLET\footnote{SCAlable supeRnet with Learnable Equivariant sTablizer} series  of several depths, especially SCARLET-A obtains 76.9\% top-1 accuracy on ImageNet. 
\end{abstract}

%%%%%%%%% BODY TEXT

\section{Introduction}

\begin{figure}[ht]
	\centering
	\includegraphics[scale=0.65]{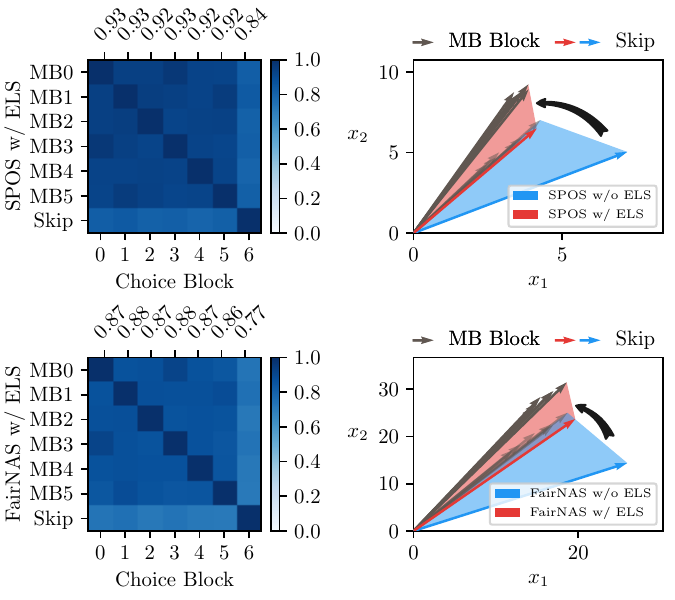}
	\caption{ELS helps to calibrate feature inconsistencies in scalable supernets with boosted cosine similarity (compared to Fig.\ref{fig:spos-fairnas-cosine-angle-3rd-no-els}) and reduced angles among feature vectors (from blue shade to red). The cosine similarity is calculated on the third layer's output feature maps from 7 paralleled choice blocks (MobileNetV2's blocks numbered 0 to 5 and a skip connection). Each block's feature vectors are projected to 2-dimensional space $(x_1, x_2)$ to draw  their relative angles (shaded in color). Other layers also have similar results. \textbf{Top:} Single Path One-Shot \protect\cite{guo2019single}, \textbf{Bottom:} FairNAS \protect\cite{chu2019fairnas}.}
	\label{fig:spos-fairnas-cosine-angle-3rd-els}
\end{figure} 

%Simplicity is the ultimate sophistication. – Leonardo da Vinci. 
Incorporating scalability into neural architecture search is crucial to exploring efficient networks. The handcrafted way of scaling models up and down is to stack more or fewer cells \cite{he2016deep,zoph2018learning}. However, model scaling is nontrivial which involves tuning width, depth, and resolution altogether. To this end, a compound scaling method is proposed in \cite{tan2019efficientnet}, it starts with a searched mobile baseline EfficientNet-B0 and `grid-search' the combination of these three factors to achieve larger models. In this paper, we are mainly concerned about finding models of varying depths, while the input resolution is kept fixed since it can be simply scaled manually.

To achieve such scalability, we first need to construct a search space of variable depths. To this end, skip connections are commonly used in differentiable approaches \cite{cai2018proxylessnas,wu2018fbnet}, but they face a common issue of undesired \emph{skip connection aggregation} as noted by \cite{chen2019progressive,zela2019understanding}, %,chu2019fairdarts
which yields non-optimal results. Recent advances in one-shot approaches take a two-stage mechanism: single-path supernet optimization and searching \cite{guo2019single,chu2019fairnas}. A supernet is an embodiment of the search space, whose single path is a candidate model. 
%,chu2019moga
Their single-path paradigm is more efficient and less error-prone, which also potentially avoids the aggregation problem but they carefully removed skip connections from search space. In this light, we integrate skip connections in their search space under the same single-path setting for a comprehensive investigation. We name the supernet in this new search space as a \emph{scalable supernet}.

Our contributions can be summarized as follows,

\textbf{First}, we are the first to thoroughly investigate scalability in one-shot neural architecture search. We discover that a vanilla training of scalable supernets suffers from instability (see Fig.~\ref{fig:spos_fairnas_train_and_histo}) and leads to weak evaluation performance. As FairNAS \cite{chu2019fairnas} suggests that feature similarity is critical for single-path training, we find that this requirement is rigorously broken by skip connections (Fig. \ref{fig:spos-fairnas-cosine-angle-3rd-no-els}).

\begin{figure*}[ht]
	\centering
	\includegraphics[scale=0.65]{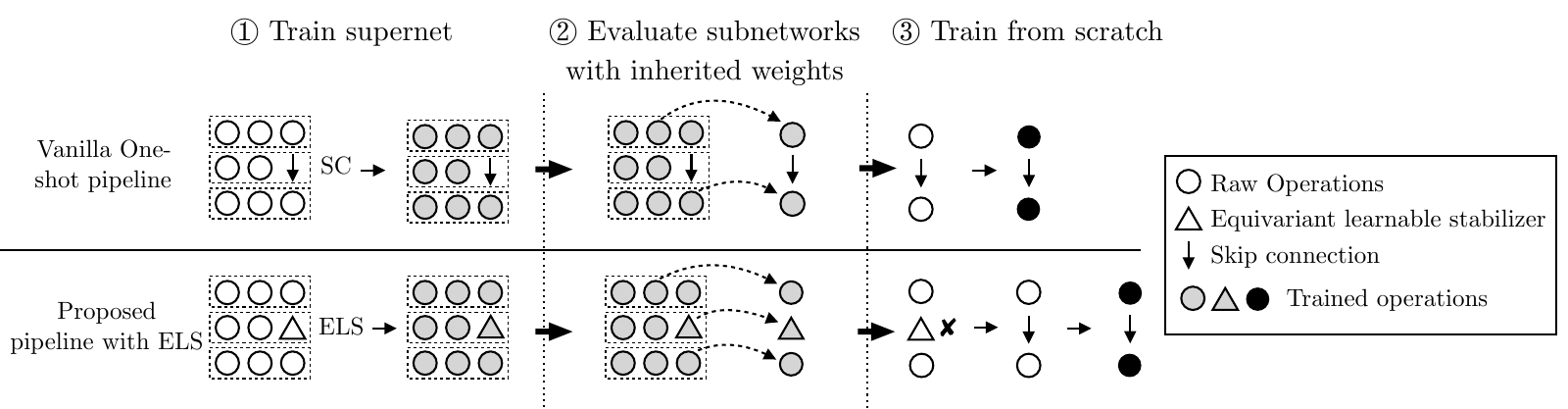}
	\vskip -0.05in
	\caption{Our proposed SCARLET-NAS pipeline where the scalable supernet is stabilized by ELS, which also provides good ranking ability compared with the vanilla approach. ELS is removed from the final subnetwork to train from scratch. Note SC and ELS can appear on each row (layer), only one is drawn for brevity.}
	\label{fig:scarlet-nas-pipeline}
\end{figure*}

\textbf{Second}, based on the above observation, we propose a simple \emph{learnable stabilizer} to calibrate feature deviation (see Fig.\ref{fig:spos-fairnas-cosine-angle-3rd-els}). It is proved effective to restore stability (see Fig.~\ref{fig:spos_fairnas_train_and_histo}) while all submodels still have invariant representational power.  Experiments on NAS-Bench-101 \cite{ying2019bench} testify that it also substantially improves the ranking performance which is crucial for the second searching stage. Our pipeline is exemplified in Fig.~\ref{fig:scarlet-nas-pipeline}.

\textbf{Last but not the least}, we perform a single proxyless evolutionary search on ImageNet after training the scalable supernet. The overall cost sums up to \textbf{10 GPU days}. Three new state-of-the-art models of different depths are generated. Specifically, SCARLET-A obtains \textbf{76.9\%} top-1 accuracy on ImageNet with 25M fewer FLOPS than EfficientNet-B0 (76.3\%)\footnote{Searching EfficientNet-B0 is similar to MnasNet \cite{tan2018mnasnet} which takes 2304 TPU days.}. Moreover, we manually upscale the searched models with \textbf{zero cost} to have  comparable FLOPS with EfficientNet variants and we also achieve competitive results.

\section{Preliminary Background}

\subsection{Single-Path Supernet Training}\label{sec:single}
The weight-sharing mechanism is now widely applied in \emph{neural architecture search} as it saves a tremendous amount of computation \cite{pham2018efficient,liu2018darts,bender2018understanding}. It is usually embodied as a supernet that incorporates all subnetworks. The supernet is trained till convergence only once, from which all subnetworks  can inherit weights for evaluation (so-called \emph{one-shot models}) without extra fine-tuning. It is thus named the \emph{one-shot} approach, as opposed to those who train each child network independently \cite{zoph2018learning,tan2018mnasnet}. Methods vary on how to train the supernet. In this paper, we concentrate on the single-path way \cite{guo2019single,chu2019fairnas}, which is more memory-friendly and efficient.

Single Path One-Shot \cite{guo2019single} utilizes a supernet $\mathcal{A}$ with 20 layers, and there are 4 choice blocks per layer based on ShuffleNet \cite{zhang2018shufflenet}. The total size of the search space reaches $4^{20}$. It uniformly samples a single-path model (say $a$ with weights $W_a$) to train at each step, after which only this activated path in the supernet gets its weights $W_a$ updated. Formally, this process is to reduce the overall training loss $\mathcal{L}_{train}$ of the supernet,

\begin{align}
W_{\mathcal{A}} = \text{argmin}_{W} \mathbb{E}_{a\sim\Gamma_{\mathcal{A}}} [ \mathcal{L}_{train} (\mathcal{A}(a, W_a))]
\end{align}

Notice that it differs from the nested manner in differential approaches \cite{liu2018darts,dong2019searching} where $\Gamma$ is not fixed but used as a representation for variable architectural weights. 

FairNAS \cite{chu2019fairnas} rephrases each supernet training step as training $m$ single-path models either sequentially or in parallel. These models are built on choice blocks \emph{uniformly sampled without replacement} (denoted as $a \sim \Psi_{\mathcal{A}}$). During each step, all blocks in the supernet are trained once. The weights are aggregated and also updated once in a single step. It can be formulated as,

\begin{align}
W_{\mathcal{A}} = \text{argmin}_{W} \mathbb{E}_{a \sim\Psi_{\mathcal{A}}} [ \frac{1}{m} \sum^m_{i}   \mathcal{L}_{train} (\mathcal{A}(a_i, W_{a_i}))]
\end{align}

%Note both methods treat the trained supernet as an evaluator. 

By ensuring the same amount of training for each block, FairNAS achieves a notable improvement in supernet performance. Interestingly enough, features learned by each block (of the same layer) in thus-trained supernet have high channel-wise similarities. This will be later proved a useful hint to restore training stability when skip connections are involved. %\footnote{Cosine similarities are above 0.92 by average.}

\subsection{Model Ranking}
Searching is essentially based on ranking. Incomplete training can give a rough guess \cite{zoph2018learning} but it is too costly. Differentiable methods \cite{liu2018darts} consider the magnitude of architectural coefficients as each operation's importance. However, there is a large discrepancy when discretizing such continuous encodings.  As we are focusing on the two-stage weight-sharing neural architecture search method, we rely on the supernet to evaluate models. It is thus of uttermost importance for it to have a good model ranking ability. FairNAS \cite{chu2019fairnas} has shown that \emph{strict fairness} during supernet training has a strong impact on it.  In particular, they adopted Kendall Tau \cite{kendall1938new} to measure the correlation between the performance of one-shot models (predicted by the supernet) and stand-alone models (trained from scratch). Tau value ranges from -1 to 1, meaning the order is completely inverted or identical. Ideally, we would like a tau of 1, which gives the exact ground truth ranking of submodels.

\section{Training Instability of Scalable Supernet}\label{sec:instability}

\subsection{Degraded Supernet Performance}

\begin{figure}[ht]
	\centering
	\includegraphics[scale=0.55]{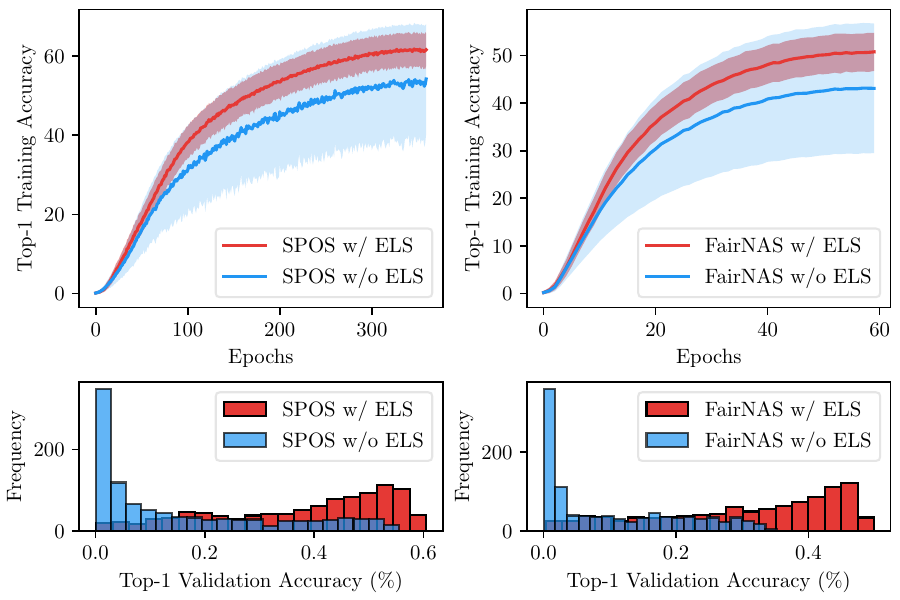}
	\caption{Training supernet with Single Path One-Shot \protect\cite{guo2019single} and FairNAS \protect\cite{chu2019fairnas} on ImageNet with and without Equivariant Learnable Stabilizer (ELS) in search space $S_1$. \textbf{Top:} The supernets with ELS enjoy better convergence (red thick lines) and small variance (red shaded area). \textbf{Bottom:} Histogram of randomly sampled 1k one-shot models'  accuracies. Supernets with ELS have an improved estimation of subnetworks.}
	\label{fig:spos_fairnas_train_and_histo}
\end{figure}

The skip connection plays a role in changing depths for architectures in MobileNetV2's block-level search space  \cite{cai2018proxylessnas,wu2018fbnet}. We detail it as $S_1$ and its variant $S_2$  in \ref{sec:ss}. %However, it is manually excluded in SPOS \cite{guo2019single} and FairNAS \cite{chu2019fairnas}. 
To investigate \emph{scalability} in one-shot approaches, we train the supernet in the previously discussed single-path fashion (Section~\ref{sec:single}) in search space $S_1$. Surprisingly, we find them suffering from severe \textbf{training instability}, which is illustrated in Fig.~\ref{fig:spos_fairnas_train_and_histo}. Unlike the reported stable training process for the supernets without skip connections \cite{guo2019single,chu2019fairnas}, we instead observe much higher variances (shadowed in blue at the top of Fig.~\ref{fig:spos_fairnas_train_and_histo}) and lower training accuracies (solid line in blue).  

Training instability also deteriorates one-shot model performance. We sample 1024 models to measure their accuracies on the ImageNet validation dataset. Fig. \ref{fig:spos_fairnas_train_and_histo} demonstrates that the majority of one-shot models from both SPOS (bottom left in blue)  and FairNAS (bottom right in blue) are underestimated, which are mainly close to 0. This phenomenon hasn't been observed in reduced $S_1$ (without skip connections) by previous work.

In particular, we need to neither overestimate nor underestimate the sampled submodels. This is hard for the scalable supernet trained so far. We can easily draw an example in Table~\ref{fig:underest-overest}, where model A is underestimated with only 1\% accuracy and B overestimated (49\%, much better than A). The ground truth is just the opposite, A has 74\% which is better than B with 73.3\%. We later show how we design an ELS for the supernet training to rectify this mistake.

\setlength{\tabcolsep}{4pt}
\begin{table}
	\centering
	%single-column
	%\begin{tabular} {|l*{3}{|r}|}
	%double-colum
	\begin{tabular}{|p{2.6cm}*{3}{|r}|}
		\hline
		  Models   & Top-1 ($\%$)  & Top-1 ($\%$) & Top-1 ($\%$)  \\
		 (in $S_1)$&(w/o ELS)&(w/ ELS)& (standalone)
		\\
		\hline
		A(0,5,0,6,3,2,3,0,2, 1,3,5,2,4,4,4,5,3,6) & 1.0  & 53.1 & 74.0\\
		\hline
		B(5,0,1,0,2,6,6,4,3, 1,5,1,0,2,4,4,1,1,2) & 49.5&49.6 & 73.3 \\
		\hline
	\end{tabular}
	\caption{ImageNet performance of model A and B (denoted by the choice block IDs) in $S_1$. Both are mistakenly estimated by the supernet trained w/o ELS. Instead, enabling ELS gives the right ranking.}
	\label{fig:underest-overest}
\end{table}

\subsection{Skip Connections Break Feature Similarity}

A well-trained supernet matters for one-shot models' ranking. We are thus driven to unveil what causes such a phenomenon to find a cure for stabilizing the training process. 

\begin{figure}[ht]
\centering
\includegraphics[scale=0.6]{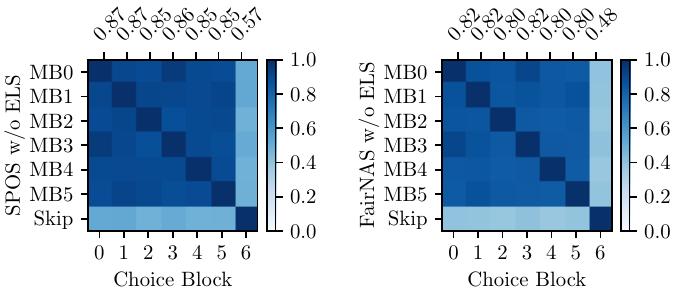}
\caption{Cosine similarity matrices of the third layer's outputs (averaged on 32 channels of $28\times28$ feature maps) from 7 choice blocks of supernets trained without ELS. The average similarity is shown as $x$-axis at the top. The skip connection yields different feature maps from others. \textbf{Left:} Single Path One-Shot \protect\cite{guo2019single}, \textbf{Right:} FairNAS \protect\cite{chu2019fairnas}}
\label{fig:spos-fairnas-cosine-angle-3rd-no-els}
\end{figure}

Inspired by the analysis of the underlying working mechanism in the single-path training \cite{chu2019fairnas}, we pick the outputs of the third layer (for an example) in the formerly trained supernets to calculate their cosine similarities across different choice blocks, which are depicted as $7\times7 $ similarity matrices in Fig.~\ref{fig:spos-fairnas-cosine-angle-3rd-no-els}.  The first six inverted bottlenecks of different configurations yield quite similar high-dimensional features (with a shape of $32\times28\times28$) and their cosine similarities are high (all above 0.85). Meanwhile, the feature maps from the skip connection (the last choice block) are quite distinct from other blocks and the average cosine similarity is below 0.6.  This disparity is observed in both training methods. 

%However, the similar feature maps of the first six choices can be regarded as a weak feature augmentation without affecting training stability. Notice even when a single layer fails, the overall supernet suffers too. 
Feature disparity troubles the training for the next layer and consequently the whole supernet. As the fourth layer randomly selects one output from the third layer, the unique skip connection disrupts feature similarities. This discrepancy of inflowing features (occurs in other layers too) will get magnified layer by layer and finally deteriorate supernet training. This is shown on the top of Fig.~\ref{fig:spos_fairnas_train_and_histo}. What's worse, it makes big trouble for the supernet to predict submodels' performance. Such a supernet becomes nearly useless because it severely underestimates or overestimates candidate architectures, shown at the bottom of Fig.~\ref{fig:spos_fairnas_train_and_histo}. Therefore, we attribute the instability to low similarities of features across different paralleled choices, mainly from skip connections. 

\section{Scalable Neural Architecture Search}
\subsection{Improve Supernet Training with a Learnable Stabilizer}
Based on the previous discussion, one direct approach to stabilize the training process is to boost the cross-block similarities by replacing the \emph{parameter-free} skip connection with a learnable stabilizer. Ideally, the stabilizer will deliver similar features as other choice blocks. What's more important, the stabilizer must be equivariant in terms of representational capacity since we want to remove it eventually (see the third step in Figure \ref{fig:scarlet-nas-pipeline}). This is detailed as Definition \ref{def:els}.

\begin{definition}\label{def:els}
\textbf{Equivariant Learnable Stabilizer.}
A plug-in learnable stabilizer is called an Equivariant Learnable Stabilizer (ELS) iff a model with such a stabilizer is exactly equivalent to the one without it in terms of representational capacity.
\end{definition}

For a search space $S$ like $S_1$ with $n$ choices per layer,  we denote $x^{c_l}_l$ as the input with $c_l$ channels to layer $l$, and $f_l^{o}$ the $o$-th operation function in that layer. Without loss of generality, we put the skip connection as the last choice, while other choices all start with a convolution operation. The equivalence requirement for an equivariant learnable stabilizer function $f_l^{ELS}$ can then be formulated as, 
\begin{equation}{\label{eq:requirment}}
	f_{l+1}^{o}(x_l^{c_l})=f_{l+1}^{o}(f_l^{ELS}(x_l^{c_l})), \forall o \in \{0,1,2,...,n-1\}.
%\begin{split}
%
%\end{split}
\end{equation}
As for $S$, we can utilize the property of matrix multiplication to find a simple ELS function:  a $1\times1$ convolution without batch normalization or activation. This is given as Lemma \ref{lem:els} and proven in the \ref{sec:app-proof}.

\begin{lemma}\label{lem:els}
	Let $f_l^{ELS}=Conv_{(c_l,c_{l+1},1,1)}$, then Equation~\ref{eq:requirment} holds. 
	
%	$ with an identity operation and another 2D convolution $Conv_{(c_{l-1},m,k,1)}$ to construct $B$ from $A$, then we can ensure $A=B$.
\end{lemma}

By adopting the learnable 1 $\times$ 1 convolution as an ELS, we observe improved stability in supernet training and better evaluation of subnetworks (Fig.~\ref{fig:spos_fairnas_train_and_histo}). We still maintain scalability since we can remove ELS based on Equation \ref{eq:requirment}.

\subsection{Neural Architecture Search with the Scalable Supernet}
Being a two-stage one-shot method like \cite{bender2018understanding,guo2019single,chu2019fairnas}, we have so far focused on supernet training. For the second searching stage, evolutionary algorithms are mostly used. For instance, FairNAS \cite{chu2019fairnas} utilizes the well-known NSGA-II algorithm \cite{deb2002fast} where they examine three objectives: classification accuracies, multiply-adds and the number of parameters. In practice, they are of different importance. We are more concerned  about accuracies (performance) and multiply-adds (speed) than the number of parameters (memory cost), which calls for a weighted solution like \cite{chu2019moga}.  It is however nontrivial for our scalable search space. First of all, models with too many skip connections are easily sorted as frontiers because of low multiply-adds. Although such a model dominates others but it usually comes with a low accuracy which is not desired. So we set a minimum accuracy requirement $acc_{min}$. Second, we are searching models for mobile deployment, where we should encourage increasing the number of parameters to prevent underfitting rather than overfitting \cite{zhang2018shufflenet}. Last, for practical reasons, we also need to set maximum multiply-adds $madds_{max}$. Formally, we describe our searching process as a constrained multi-objective optimization as follows,

\begin{equation}
\label{eq:three_objs}
\begin{split}
max & \quad \{ acc(m), -madds(m), params(m)\}, \\
\quad m &\in \enskip \text{search  space} \enskip  S\\
s.t. &\quad w_{acc} + w_{madds}  + w_{params} = 1, \forall w >=0 \\
& \quad acc(m) > acc_{min}, madds(m) < madds_{max}.
\end{split}
\end{equation}

Specifically, we adopt a similar evolutionary searching algorithm based on NSGA-II \cite{deb2002fast} as in FairNAS \cite{chu2019fairnas} with some modifications. For handling weights of different objectives, we make use of weighted crowding distance \cite{friedrich2011weighted} for non-dominated sorting. We set $w_{acc}=0.4, w_{madds}=0.4, w_{params}=0.2$. The constraints are set to $madds_{max}=500M$ and $acc_{min} = 0.4$. Notice that we treat these two constraints in sequential order to reduce cost. As calculating multiply-adds is much faster than accuracies, models violating $madds_{max}$ are immediately removed for further evaluation. The whole search pipeline is presented in Algorithm~\ref{alg:nas_pipeline} and Fig. \ref{fig:nas-pipeline} (both in \ref{sec:app-alg}). 

\section{Experiments}
\subsection{Dataset, Training, and Searching}
\textbf{Dataset.} For training and searching, we use the ILSVRC2012 dataset \cite{deng2009imagenet}. To be consistent with previous work \cite{tan2018mnasnet}, the validation set consists of 50k images selected from the training set. The original validation set serves as the test set. 

\textbf{Supernet Training.} For FairNAS experiments in $S_1$, we follow \cite{chu2019fairnas} except that we train the supernet for 60 epochs. It costs nearly 8 GPU days. For SPOS, we train it for 360 epochs to have the same amount of weight updates per block. As for $S_2$ with more choices, we use the same setting except for a smaller batch size of 256, which results in higher top-1 accuracy on average. % to have more back-propagation iterations per epoch

\begin{figure*}[ht]
	\centering
	\includegraphics[scale=0.25]{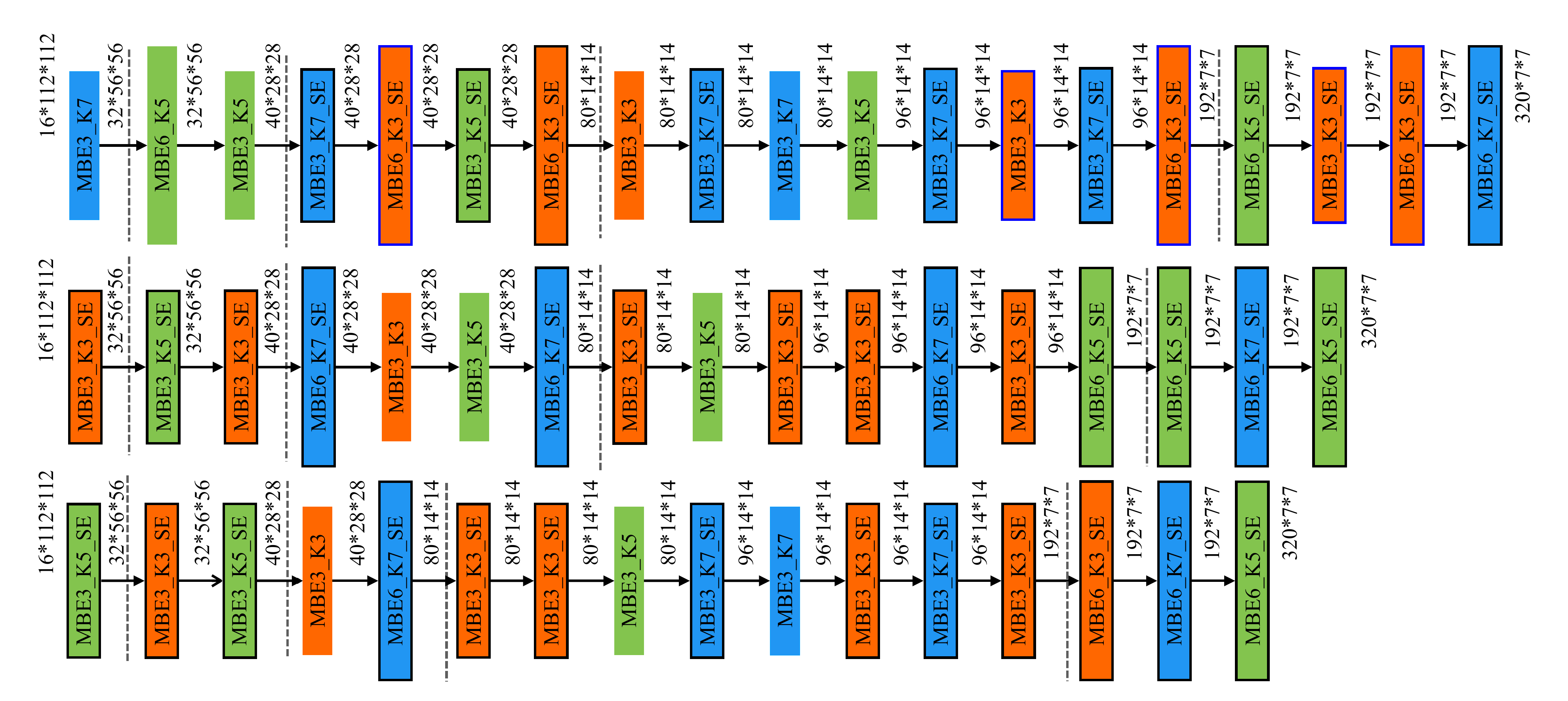}
	%\vskip -0.1in
	\caption{The architectures of SCARLET-A, B and C (from top to bottom). Downsampling points are indicated by dashed lines. The stem and tail parts are omitted for brevity.}
	\label{fig:scarlet-models}
	%\vskip -0.1in
\end{figure*}

\textbf{Evolutionary Searching.} We search proxylessly in $S_2$ on ImageNet. The evolution covered 8400 models (a population of 70 models evolved for 120 generations). It costs 2 GPU days on a Tesla V100. The final architectures SCARLET-A, B and C (shown in Fig. \ref{fig:scarlet-models}) are sampled from the Pareto front at equal distance and are trained from scratch. Due to the equivalence requirement, we remove ELS to achieve two competitive models with shorter depths, SCARLET-B and C. 

\textbf{Single Model Training.} To train the selected single model,  we follow MnasNet \cite{tan2018mnasnet} with vanilla Inception preprocessing \cite{szegedy2017inception}. We train EfficientNet and SCARLET models without AutoAugment \cite{cubuk2018autoaugment} to have a fair comparison with state-of-the-art architectures. The batch size is 4096. The initial learning rate is 0.256 and it decays at an amount of 0.01 every 2.4 epochs. The dropout with a rate 0.2 \cite{srivastava2014dropout} is put before the last FC layer. The weight decay rate ($l_2$) is $1.0\times10^{-5}$. The RMSProp optimizer has a momentum of 0.9.

\subsection{ImageNet Classification}

\setlength{\tabcolsep}{4pt}
\begin{table}
\centering
\begin{tabular}{|l|r|r|r|r|}  
\hline
Models  & $\times$+  & Params  & Top-1 & Top-5 \\
&(M)&(M)&($\%$)&($\%$)
\\
\hline
MobileNetV2 \cite{sandler2018mobilenetv2}       & 300  & 3.4 & 72.0 & 91.0     \\
MobileNetV3 \cite{howard2019searching} & 219 & 5.4  &75.2 & 92.2   \\
MnasNet-A1 \cite{tan2018mnasnet} & 312 & 3.9 & 75.2 & 92.5   \\
MnasNet-A2 \cite{tan2018mnasnet}  & 340 & 4.8 & 75.6 & 92.7 \\
FBNet-B \cite{wu2018fbnet} & 295 & 4.5 & 74.1 & -\\ 
Proxyless-R \cite{cai2018proxylessnas}   & 320 & 4.0 & 74.6 & 92.2 \\
Proxyless GPU \cite{cai2018proxylessnas}  & 465 & 7.1  & 75.1 & -\\
Single-Path \cite{stamoulis2019single} & 365 & 4.3 & 75.0 & 92.2 \\ 
SPOS \cite{guo2019single}  & 328 & 3.4 &74.9 & 92.0 \\
FairNAS-A \cite{chu2019fairnas} & 392& 5.9& 77.5 & 93.7 \\
FairDARTS-C \cite{chu2019fair}& 386 & 5.3 & 77.2 & 93.5 \\
DARTS- \cite{chu2020darts} & 470  & 5.5 &  77.8& 93.9   \\
%FairNAS-A \cite{chu2019fairnas}  & 388& 4.6 &  75.3 & 92.4 \\ %(Xiaomi) 
%FairDARTS-B \cite{chu2019fair}& 541 & 4.8 &75.1 & 92.5 \\
%DARTS- \cite{chu2020darts} & 467 & 4.9 & 76.2 & 93.0  \\
MixNet-M \cite{tan2020mixconv} & 360 & 5.0 & 76.6$^\dagger$ (77) & 93.2 \\
EfficientNet B0 \cite{tan2019efficientnet}  & 390 & 5.3 &  76.3 & 93.2 \\
SCARLET-A (Ours) & 365 & 6.7 & \textbf{76.9} & \textbf{93.4}  \\
SCARLET-B (Ours) & 329 & 6.5 &76.3 & 93.0  \\
SCARLET-C (Ours) & 280 & 6.0 & 75.6 & 92.6  \\
\hline
\end{tabular}
\smallskip
\caption{Comparison with state-of-the-art architectures on ImageNet classification task.$^\ddagger$: model trained from scratch by us without AutoAugment.}
\label{tab:comparison-imagenet}
\end{table}

\subsubsection{Comparison of State-of-the-art Mobile Architectures}

We give full train results of the SCARLET series on ImageNet dataset in Table~\ref{tab:comparison-imagenet}. Although in absence of AutoAugment tricks \cite{cubuk2018autoaugment}, SCARLET-A still clearly surpasses EfficientNet-B0  ($+0.6\%$ higher accuracy)  using fewer FLOPS.  The shallower model SCARLET-B achieves $76.3\%$ top-1 accuracy with 329M FLOPS, which exceeds several models of a similar size by a clear margin: MnasNet-A1 ($+1.1\%$), Proxyless-R ($+1.7\%$). Notably, to be comparable to our shallowest model SCARLET-C ($75.6\%$), MnasNet-A1 comes with $21\%$ more FLOPS at the cost of $200\times$ GPU days. Even without mixed convolution, SCARLET-A  still outperforms MixNet-M \cite{tan2020mixconv}, which has $76.6\%$ accuracy when we trained it with the same tricks.
We further give a closer examination of the SCARLET series in \ref{sec:analysis-scarlet}.

\subsubsection{Comparison of Models at a Larger Scale}
Higher accuracy requirements beyond mobile settings are also considered. To be comparable with EfficientNet's scaled variants, we simply \emph{manually upscale} our SCARLET baseline models to have the same resolution and FLOPS without any extra tuning cost. We compare the results with other state-of-the-art methods in Table~\ref{tab:scaling}. 

\setlength{\tabcolsep}{4pt}
\begin{table*}
	\begin{center}
	\begin{small}
			\begin{tabular}{*{8}{|l}|} 		
				\hline
				Methods & Resolution & Depth& Channel & $\times$+  & Params & Top-1  & Top-5  \\
				&  & ($\times$) &  ($\times$) & (B) & (M) & (\%) & (\%) \\
				\hline
				DenseNet-264 \cite{huang2017densely} &224$\times$224&-&-& 6 & 34&77.9&93.9 \\
				Xception \cite{chollet2017xception} &299$\times299$&-&-& 8.4 & 23 & 79.0 & 94.5 \\
				EfficientNet B2 \cite{tan2019efficientnet}& 260$\times$260 & 1.2 & 1.1 & 1.0  & 9.2 & 79.4$^*$ & 94.7$^*$ \\ % latest 6.9
				\textbf{SCARLET-A2}$^{\dagger}$  &260$\times$260 &1.0 &1.4& 1.0  & 12.5 & 79.5  & 94.8
				\\ % latest 6.9 %
				\hline
				ResNeXt-101 \cite{xie2017aggregated} & 320$\times320$ &-&-&32&84 & 80.9 & 95.6 \\
				PolyNet \cite{zhang2017polynet}  &331$\times331$ &-&-& 35&92&81.3&95.8 \\
				SENet \cite{hu2018squeeze}
				&320$\times320$&-&-&42 & 146 & 82.7 & 96.2 \\
				EfficientNet B4 \cite{tan2019efficientnet} &380$\times$380 &1.8& 1.4 & 4.2  & 19 & 82.6 & 96.3 \\
				\textbf{SCARLET-A4}$^{\dagger}$  & 380$\times$380 & 2.0& 1.4& 4.2  & 27.8 & 82.3  & 96.0 \\
				\hline
			\end{tabular}
	\end{small}
	\end{center}
	\caption{Single-crop results of scaled architectures on ImageNet validation set. $^*$: Retrained without fixed AutoAugment (AA),$^{\dagger}$: w/o fixed AA.}
	\label{tab:scaling}
\end{table*}

At the level of 1 billion FLOPS, while EfficientNet-B2 is based on grid search at a very high cost on GPUs \cite{tan2019efficientnet}, our SCARLET-A2 (79.5\%) is from upscaling with zero cost. No AutoAugment tricks are applied for a fair comparison.  Moreover, Xception \cite{chollet2017xception} uses 8 times more FLOPS to reach $79.0\%$.  Notably, our SCARLET-A4 achieves new state-of-the-art top-1 accuracy \textbf{82.3\%} again without extra costs using only 4.2B FLOPS. By contrast, SENet \cite{hu2018squeeze} uses 10$\times$.

\subsection{Transferability to CIFAR-10}

Table~\ref{table:scarlet-transfer-cifar10} shows our transfer results on CIFAR-10 dataset \cite{krizhevsky2009learning}. We utilize similar training settings from \cite{kornblith2019better}. In particular, each model is loaded with ImageNet pre-trained weights and finetuned for 200 epochs with a batch size of 128. The initial learning rate is set to 0.025 with a cosine decay strategy. We also adopted AutoAugment policy for CIFAR-10 \cite{cubuk2018autoaugment}. The dropout rate is 0.3. To achieve comparable top-1 accuracy as NASNet-A Large, our SCARLET-A only uses $\mathbf{33\times}$ fewer FLOPS. SCARLET-B doesn't utilize the mixed convolution but it is still comparable to MixNet \cite{tan2020mixconv}. In particular, our smallest model SCARLET-C is close to MixNet-M, saving 22$\%$ FLOPS.

\begin{table}[h]
\begin{center}
\begin{tabular}{|l|c|r|Hl|}
\hline
Models & Input Size & $\times+ $ (M) & Params (M) & Top-1 (\%)\\
\hline
%NASNet-A \cite{zoph2018learning}$^\dagger$ & 12030 & 85 & 98.0 \\
%MixNet-M \cite{tan2020mixconv} & 352 & 3.5 & 97.92 \\
%SCARLET-A & 389 & 5.4 & 98.05\\
%SCARLET-B & 348 & 5.3 & 97.93\\
%SCARLET-C & 299 & 4.8 & 97.91\\
%=======
NASNet-A Large \cite{zoph2018learning}$^\dagger$ & 331$\times$331 & 12030 & 85 & 98.00\\
MixNet-M \cite{tan2020mixconv} &224$\times$224&  352 & 3.5& 97.92\\
SCARLET-A  &224$\times$224& 364 & 5.4 & \textbf{98.05}\\
SCARLET-B  &224$\times$224& 328 & 5.3 & 97.93\\
SCARLET-C  &224$\times$224& 279 & 4.8 & 97.91\\
%>>>>>>> elaborate cifar10 transfer.
\hline
\end{tabular}
\end{center}
\caption{Transferring SCARLET models to CIFAR-10. $^\dagger$: Reported by \cite{kornblith2019better}.}
\label{table:scarlet-transfer-cifar10}
\end{table}

%Table~\ref{table:scarlet-transfer-cifar10} shows our transfer results on CIFAR-10. Each model is trained for 200 epochs with a batch size of 128. The initial learning rate is set to 0.025 with cosine decay strategy. We also adopted AutoAugment policy for CIFAR-10 \cite{cubuk2018autoaugment}. The dropout rate is 0.3 and drop path is 0.3.

\subsection{Object Detection}

To verify the transferability of our models on the object detection task, we utilize drop-in replacements of backbones of the RetinaNet framework (Res101+FPN) \cite{lin2017focal}. To make fair comparisons, we focus on the mobile settings of the backbone. All methods are trained for 12 epochs on COCO dataset \cite{lin2014microsoft} with a batch size of 16 (\texttt{train2017} for training and \texttt{val2017} for reporting results). The initial learning rate is 0.01 and decayed by 0.1 on epoch 8 and 11. 
Compared with recent NAS models in Table~\ref{table:scarlet-coco-retina}, we utilize fewer FLOPS to have better results, suggesting a better transferability.   
%Even without the MixConv operations, our models are comparable with MixNet \cite{tan2020mixconv}.

%\setlength{\tabcolsep}{4pt}
\begin{table*}
	\begin{center}
		\begin{tabular}{*{2}{|l}|H*{7}{l|}}
			\hline
			Backbones & $\times +$  & Params &Acc    & AP & AP$_{50}$ & AP$_{75}$ & AP$_S$ & AP$_M$ & AP$_L$ \\
			%\begin{scriptsize}
			& (M) & (M) & (\%) &(\%) & (\%)& (\%)&(\%) &(\%) &(\%) %\end{scriptsize}
			\\
			\hline
			MnasNet-A2 \cite{tan2018mnasnet} & 340& 4.8 & 75.6 & 30.5 & 50.2 & 32.0 & 16.6 & 34.1 & 41.1\\
			MobileNetV3 \cite{howard2019searching} & 219 & 5.4 & 75.2& 29.9 & 49.3 & 30.8 & 14.9 & 33.3 & 41.1\\

			%			EfficientNetB0 \cite{tan2019efficientnet} & 390 & 5.3&76.3& \\
			SingPath NAS \cite{stamoulis2019single} & 365 & 4.3 & 75.0 & 30.7 & 49.8 & 32.2 & 15.4 &33.9 & 41.6\\
%			MixNet-M \cite{tan2020mixconv} & 360 & 5.0 & 77.0 & 31.3& 51.7 & 32.4& 17.0 & 35.0 & 41.9   \\
			SCARLET-A &365&6.7&76.9&\textbf{31.4}& \textbf{51.2}& \textbf{33.0}& 16.3& \textbf{35.1}&41.8\\
			SCARLET-B &329&6.5&76.3&31.2 & 51.2& 32.6 & \textbf{17.0} & 34.7& \textbf{41.9}\\
			%SCARLET-C \\
			\hline
		\end{tabular}
		\smallskip
		\caption{Object detection result of various drop-in backbones on the COCO dataset.}
		\label{table:scarlet-coco-retina}
	\end{center}
\end{table*}
%

%\subsection{Image Segmentation}

%\subsection{Domain Transfer}
\section{Ablation Study and Analysis}

%For experiment (i) and (ii), we can see that ELS with BN post calibration can boost Kendall Taus in each case. 

\subsection{Training Stability}

Compared with skip connection, ELS can help stabilize the training process of a scalable supernet, shown in Fig.~\ref{fig:spos_fairnas_train_and_histo}. We believe it is due to boosted cross-block features similarities (increased by 0.3 compared with pure skip connection). Interestingly enough, ELS is also able to close up the feature angle discrepancy. This phenomenon is depicted in Fig.~\ref{fig:spos-fairnas-cosine-angle-3rd-els}. Informally, ELS plays an important role in rectifying the features' \emph{phase gap} between skip connections and other \emph{homogeneous} choices. Essentially, ELS is a near-homogeneous  to an inverted bottleneck block, while a skip connection is instead \emph{heterogeneous}.  As a result, for both one-shot approaches, supernets with ELS enjoy higher training accuracies (red solid line) and lower variances (red shaded area). Although there is a small proportion of one-shot models with low accuracy, they can be easily excluded from the proposed constrained optimization process.

\subsection{Ranking Ability with and without ELS}

\begin{figure}
	\centering
	\includegraphics[scale=0.5]{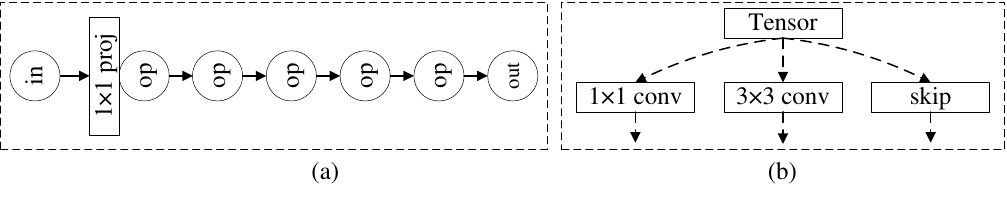}
	\caption{Ranking analysis is based on a subspace of NAS-Bench-101. \textbf{(a)} A cell is a stack of 5 nodes. An additional $1\times 1$ conv projection is added before the first one to avoid channel mismatch. \textbf{(b)} For each node, we can select one operation from 3 choices.}
	\label{fig:nas-101}
\end{figure}

The most important role of the supernet in the two-stage approach is to evaluate the performance of the subnetworks, so-called  `ranking ability'. To find out the contribution of ELS, we perform experiments on a common NAS benchmark NAS-Bench-101 \cite{ying2019bench} with some adaptations. Specifically, we construct a supernet to have a stack of 9 cells, each cell has at most 5 sequential internal nodes, each node has 3 optional operations: 1 $\times$ 1 Conv, 3 $\times$ 3 Conv and a skip connection. The first node is preceded by a 1x1 Conv projection. The designed cell and node choices are shown in Fig.~\ref{fig:nas-101}. 

\begin{figure}
	\centering
	\includegraphics[scale=0.35]{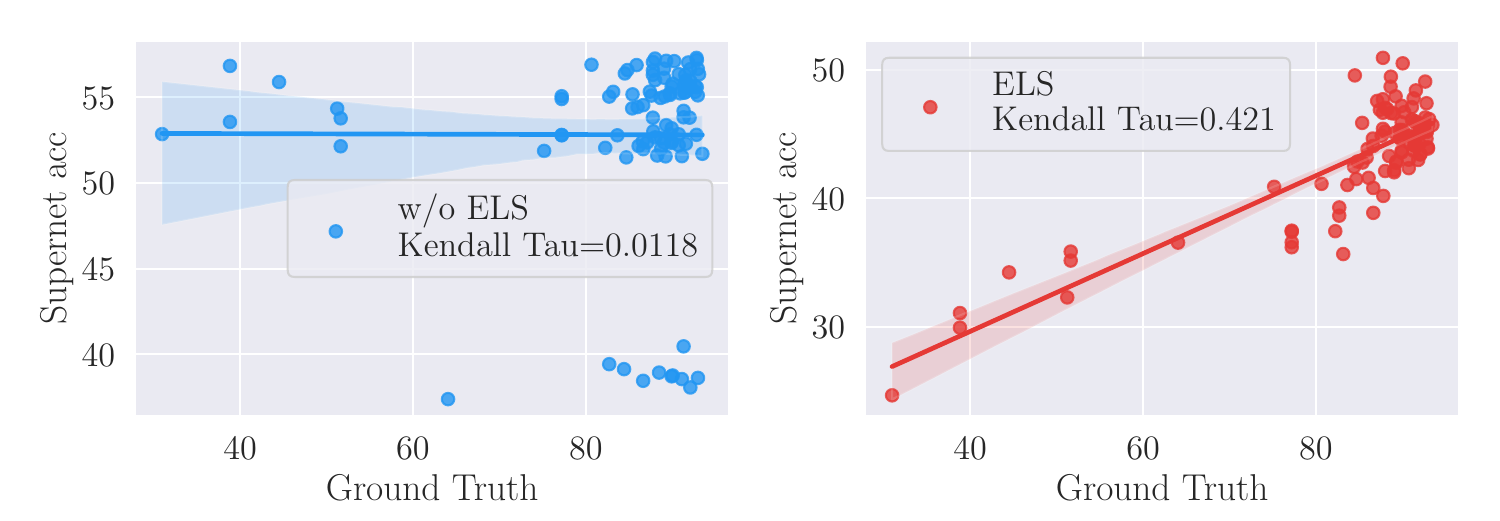}
	\caption{Comparison of ground-truth vs. estimated accuracy correlation between the supernets trained with and without ELS, based on 100 sampled models from NAS-Bench-101 \cite{ying2019bench}. ELS substantially boosts the ranking ability of the supernet.}
	\label{fig:ranking}
\end{figure} 

We train such a supernet with and without ELS on CIFAR-10. 
For both experiments, we train for 100 epochs with a batch size of 96, and a learning rate of 0.025. We randomly sample 100 models to lookup their ground-truth accuracies from NAS-Bench-101. We calculate the ranking ability of each supernet using Kendall Tau \cite{kendall1938new}, shown in Fig~\ref{fig:ranking}. The one with ELS reaches a  tau value of 0.421, indicating much higher correlation.
%Table~\ref{table:nas101-ranking}.

%\setlength{\tabcolsep}{4pt}
%\begin{table}
%	\caption{Comparison of Kendall Taus between the supernets trained with/without ELS, based on 100 sampled models from NAS-Bench-101.}
%	\centering
%	\begin{tabular} {|l*{4}{|r}|}%{p{8cm}*{3}{r}} %
%		\hline
%		Strategy   & w/o ELS$^\dagger$   & w/ ELS \\
%		\hline
%		SPOS  & 0.0118 &  0.421\\
%		FairNAS  &  &  \\
%		\hline
%	\end{tabular}
%	\label{table:nas101-ranking}
%\end{table}
%

\textbf{ELS vs. Skip Connection.} To give a clearer comparison between the skip connection (SC) and the proposed ELS, we illustrate their functionality in Table~\ref{tab:comp-sc-els}. Both operations are foldable, meaning they are used in supernet training, but later removed (folded) to build the corresponding subnetworks as they both creates an identity transformation before folding and after (see also Fig.~\ref{fig:scarlet-nas-pipeline}). The difference is that, ELS is learnable so that it gives more consistent feature maps in each layer. This is crucial to improve the supernet ranking, attested by Fig~\ref{fig:ranking}.

\setlength{\tabcolsep}{4pt}
\begin{table}
	\centering
	\begin{tabular} {|l*{5}{|c}|}
		\hline
		Op  & Foldable & Identity  & Learnable & Similarity & Ranking \\
		\hline
		SC  & \cmark & \cmark & \xmark & Low & Poor\\
		ELS  & \cmark & \cmark & \cmark & High & Good\\
		\hline
	\end{tabular}
	\caption{Comparison of Skip Connections (SC) and ELS as per foldability and ranking.} %Feature similarity is from Fig.~\ref{fig:spos-fairnas-cosine-angle-3rd-els} and Fig.~\ref{fig:spos-fairnas-cosine-angle-3rd-no-els}.
	\label{tab:comp-sc-els}
\end{table}

\subsection{Equivariant vs. Non-equivariant Stabilizer}

The equivalence requirement for the stabilizer (Equation \ref{eq:requirment}) plays a pivotal role in our approach. We evaluate a subnetwork with ELS as it is an identical proxy to the one without it. A stabilizer that violates the equivalence requirement will give wrong evaluation. 

\begin{figure}[ht]
	\centering
	\includegraphics[scale=0.65]{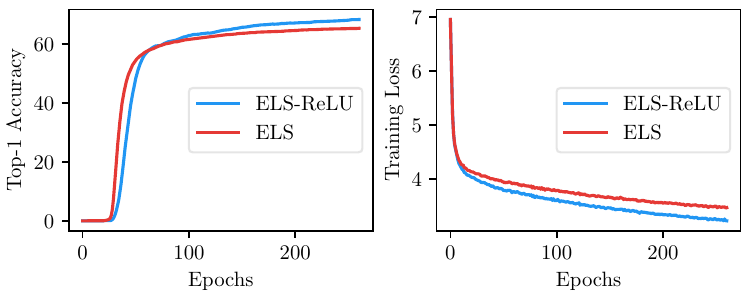}
	\caption{Adding ReLU to ELS gives a wrong evaluation (overestimation) of a subnetwork.}
	\label{fig:activation-comp}
	%\vskip -0.1in
\end{figure}

For example, we make a simple modification by adding a ReLU function to ELS, this makes the stabilizer non-equivariant because of non-linearity. Can we use a supernet with this stabilizer to correctly evaluate a model? Given a model denoted by choice indices: $M_{(1, 3, 1, 0, 12, 0, 0, 0, 12, 12, 12, 12, 12, 0, 0, 0, 12, 12, 9)}$,  when we train it respectively with ELS and with ELS-ReLU on the same settings, we find them have different representational power, as shown in Fig.~\ref{fig:activation-comp}. The one with ELS-ReLU overestimates the model compared to the one with ELS-no-ReLU, which reflects its truth by Lemma \ref{lem:els}. 

\subsection{Constrained Optimization}

For multi-objective evolutionary search in scalable search space, to limit the minimum accuracy is more than necessary. In a standard NSGA-II \cite{deb2002fast} process, models with many skip connections are easily picked for the high-ranking non-dominated sets. For an extreme example, a model consisting of skip connections for all 19 layers has minimum multiply-adds, it will always stay as a boundary node. This brings in \emph{gene contamination} for the evolution process as this poor-performing gene never dies out.

\begin{figure}
\centering
\subfigure{
\includegraphics[scale=0.7]{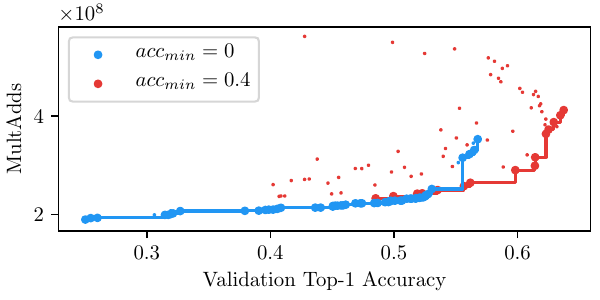}
}
%\vskip -0.05in
\subfigure{
\includegraphics[scale=0.7]{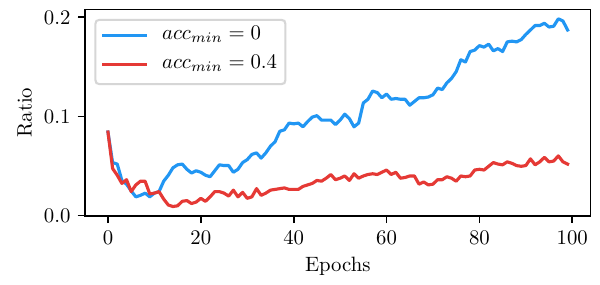}
}
\caption{Ablation study on constrained optimization. \textbf{Left:} Pareto front of MultAdds vs. Accuracy.  \textbf{Right:} The ratios of skip connections per epoch.} %The minimum accuracy constraint $acc_{min}=0.4$ not only prohibits too many skip connections' occurrences (red line at the bottom figure) in genes, but also pushes Pareto front forward (red line at the top figure).
\label{fig:ablation-constraint}
%\vskip -0.1in\dfrac{num}{den}
\end{figure}

To demonstrate the necessity of a minimum accuracy constraint, we compare the case with $acc_{min}=0$ and $acc_{min}=0.4$ in Fig.~\ref{fig:ablation-constraint}. We can observe the number of skip connections has been greatly reduced (red line in the right figure). As a consequence, the evolution converges to a better Pareto Front (red line in the left figure): higher validation accuracies at the same level of multiply-adds.

\subsection{Component Analysis}
A supernet trained without ELS can't deliver good search results. Using the same searching strategy NSGA-II, its best model found below 380M FLOPS obtains $71.3\%$ top-1 accuracy on ImageNet, while SCARLET-A (365M) has 76.9$\%$. Therefore, the contribution to the final performance comes mainly from ELS in the first stage. The searching strategy heavily depends on  ranking ability. 
%\subsection{Visualization of Equivariant Learnable Stabilizer}
%
%To further understand what does an equivariant learnable stabilizer look like, we illustrate the weights of 1$\times$1 convolutions from four layers at various depths (with equal input and output sizes) in Fig.~\ref{fig:els-weights}. Interestingly, only the diagonal weights are activated with higher values, which mimics an \emph{identity mapping}. This can be explained. As the parameter-bearing parts of inverted bottlenecks in the same layer learn residual information, the added stabilizer should capture an identity mapping as well. It also coincides with the discovered channel-wise similarity, the stabilizer only needs to deal with discrepancy channel by channel, which is very efficient.
%
%\begin{figure}
%\centering
%\includegraphics[scale=0.55]{conv_map}
%\caption{The visualization of ELS's weights in different layers. The weights are mainly activated diagonally.}
%\label{fig:els-weights}
%%\vskip -0.1in
%\end{figure}

\section{Conclusion}
In this paper, we expose a critical failure in single-path one-shot neural architecture search when scalability is considered. We discover the underlying feature dissimilarity hinders supernet training. The proposed \emph{equivariant learnable stabilizer} is effective to rectify such discrepancy while maintaining the same representational power for subnetworks. We also employ a weighted multi-objective evolutionary search to find a series of state-of-the-art SCARLET architectures. Good transferability is achieved on various vision tasks. Compared with unnecessarily costly EfficientNet, our method is a step forward towards more efficient and flexible neural architecture search. 

{\small
\bibliographystyle{ieee_fullname}
\bibliography{aaai}
}

\newpage

\appendix

\section{Proof}\label{sec:app-proof}

\begin{proof}
	%	First, we copy $B$'s weights from $A$ except for the $F_l$ and $Conv_{(c_{l},m,k,1)}$. The only thing to prove is that we can replace $F_l$ and $Conv_{(c_l,m,k,1)}$ with $Conv_{(c_{l-1},m,k,1)}$ equivalently. 
	
	First, we prove that Equation~\ref{eq:requirment} (main text) holds for $\forall o \in \{0, 1, ..., n-2\}$. In this case, it's sufficient to prove the output of the first convolution $Conv_{({c_l}, m, k,k)}$ can be exactly matched by adding $Conv_{(c_l,c_{l+1},1,1)}$ before $Conv_{(c_{l+1},m,k,k)}$.
	Let $W^1_{c_{l}, c_{l+1},1,1}$ and $W^2_{c_{l}, m,k,k}$ be the weight tensors of $Conv_{(c_l,c_{l+1},1,1)}$ and $Conv_{(c_{l+1}, m,k,1)}$ respectively. Let $W^3_{c_{l}, m,k, k}$ be the weight tensors of $Conv_{(c_{l},m,k,1)}$. Let $w$ be one element of the tensor. We have
	
	%	 any $x \in \mathbf{R}^{h,w,c_{l-1}}$, the above declaration holds. Let $W^1_{c_{l-1}, c_l,1,1}$ and $W^2_{c_{l}, m,k,k}$ be the weight tensor of $F_l$ and $Conv_{(c_l,m,k,1)}$. Let $W^3_{c_{l-1}, m,k, k}$ be the weight tensor of $Conv_{(c_{l-1},m,k,1)}$. Let $w$ be  one element of the tensor.
	\begin{gather}
	y=Conv_{(c_l,c_{l+1},1,1)}(x_l^{c_l}), z=Conv_{(c_{l+1},m,k,1)}(y)\\
	y(i,j,c)= \sum_{p=1}^{c_{l}} w^1_{p, c,1,1}  x(i,j,p) 
	\end{gather}
	Also,
	\begin{equation}
	\begin{split}
	z(i,j,c)&=\sum_{q=1}^{k}\sum_{p=1}^{c_{l+1}}w^2_{p, c,q, q}  y(i+q,j+q,p)\\
	&=\sum_{q=1}^{k}\sum_{p=1}^{c_{l+1}}w^2_{p, c,q, q}  (\sum_{u=1}^{c_{l}} w^1_{u, p,1,1}  x(i+q,j+q,u)) \\
	&=\sum_{q=1}^{k}\sum_{p=1}^{c_{l+1}} \sum_{u=1}^{c_{l}} w^2_{p, c,q, q} 
	w^1_{u, p,1,1}  x(i+q,j+q,u)\\
	%	&=\sum_{q=1}^{k} \sum_{u=1}^{c_{l-1}} \sum_{p=1}^{c_l}w^2_{p, c,q, q} 
	%	w^1_{u, p,1,1}  x(i+q,j+q,u) \\
	&=\sum_{q=1}^{k} \sum_{u=1}^{c_{l}} w^3_{u, c, q, q}  x(i+q,j+q,u)	
	\end{split}
	\end{equation}
	%	\begin{equation*}
	%	\begin{split}
	%	&=\sum_{q=1}^{k}\sum_{p=1}^{c_{l+1}} \sum_{u=1}^{c_{l}} w^2_{p, c,q, q} 
	%	w^1_{u, p,1,1}  x(i+q,j+q,u)\\
	%%	&=\sum_{q=1}^{k} \sum_{u=1}^{c_{l-1}} \sum_{p=1}^{c_l}w^2_{p, c,q, q} 
	%%	w^1_{u, p,1,1}  x(i+q,j+q,u) \\
	%	&=\sum_{q=1}^{k} \sum_{u=1}^{c_{l}} w^3_{u, c, q, q}  x(i+q,j+q,u)
	%	\end{split}
	%	\end{equation*}
	Therefore, the first part is proved by setting
	\begin{equation}
	w^3_{u, c, q, q} = \sum_{p=1}^{c_{l+1}}w^2_{p, c,q, q} 
	w^1_{u, p,1,1}. 
	\end{equation}	
	For $o=n-1$, we replace a skip connection with an $ELS$. We can iteratively apply the first part of the proof till the end of searchable layers. 
\end{proof}

\section{Algorithm} \label{sec:app-alg}

\begin{figure}[ht]
	\centering
	\subfigure{
		\includegraphics[scale=0.3]{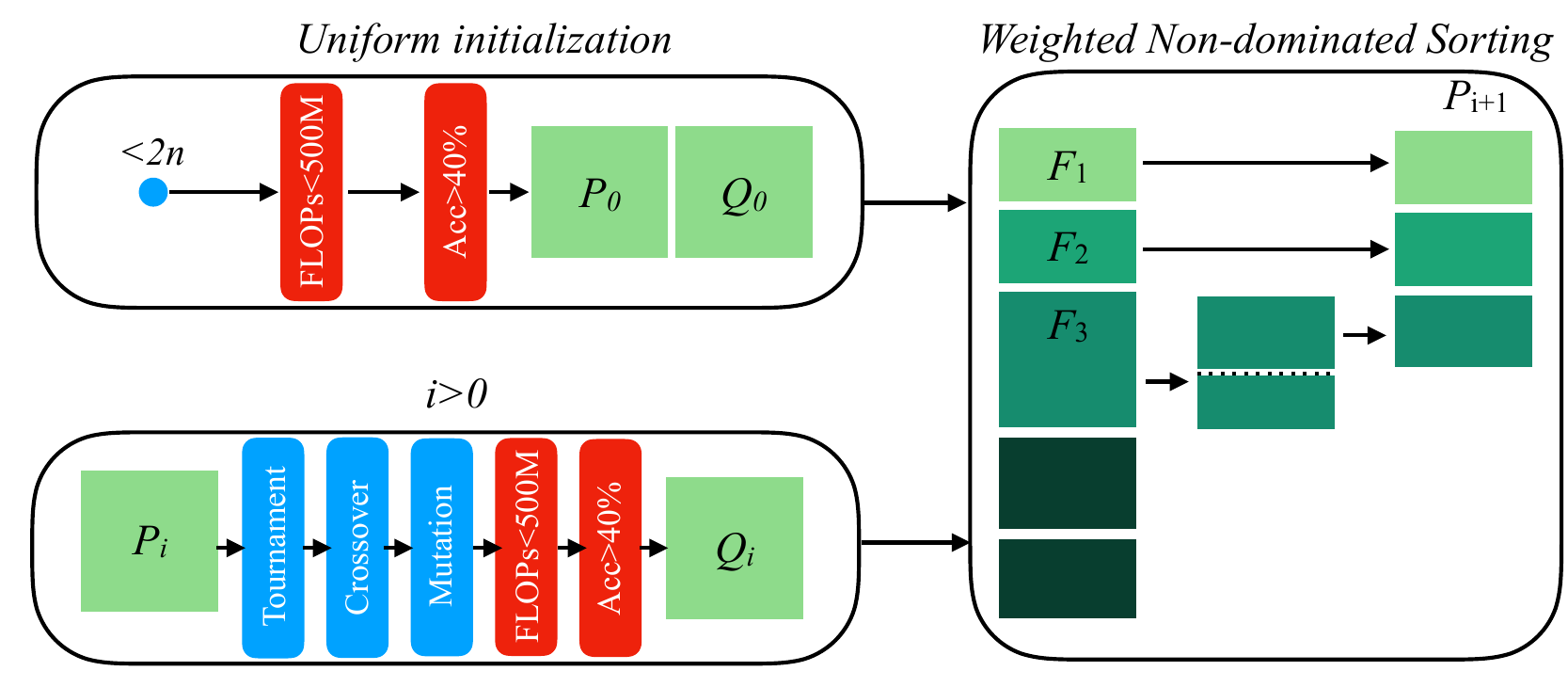}
	}
	%\vskip -0.1in
	\caption{Constrained and weighted NSGA-II Pipeline. It starts with a uniform initialization (top left) with some constraints (red) to generate the initial population. The trained scalable supernet serves as a fast evaluator to decide the relative performance of each model so that they can be grouped into several Fronts ($F_1, F_2, \ldots$) by weighted non-dominated sorting (right). Only the top $n$ of them make up the next generation $P_{i+1}$, based on which $Q_{i+1}$ is produced with tournament selection, crossover and mutation (blue) under the same constraints (bottom left). The whole evolution loops until we reach Pareto-optimality.}
	\label{fig:nas-pipeline}
	%\vskip -0.1in
\end{figure}

Our constrained and weighted NAS pipeline is listed in Algorithm \ref{alg:nas_pipeline} and Fig. \ref{fig:nas-pipeline}.

\begin{algorithm}[tb]
	\caption{The constrained and weighted NAS pipeline.}
	\label{alg:nas_pipeline}
	\begin{algorithmic}
		\STATE {\bfseries Input:} Supernet $S$, the number of generations $N$, population size $n$, validation dataset $D$, constraints $C$, objective weights $w$
		\STATE {\bfseries Output: } A set of $K$ individuals on the Pareto front.
		\STATE {Train supernet $S$ defined on the scalable search space.}
		
		\STATE {Uniformly generate the populations $P_0$ and $Q_0$ until each has $n$ individuals satisfying $C_{\text{FLOPS}}$, $C_{\text{Accuracy}}$.}
		\FOR {$i=0$ {\bfseries to} $N-1$}
		\STATE{$R_i = P_i \cup Q_i$}
		\STATE {$F = \text{non-dominated-sorting}(R_i)$}
		%			\STATE {$L = \text{improved-crowding-distance}(F)$}
		\STATE{Pick $n$ individuals to form $P_{i+1}$ by ranks and the crowding distance \textbf{weighted} by $w$.}
		\STATE {$Q_{i+1} = \emptyset$}
		\WHILE {$size(Q_{i+1}) < n $}

		\STATE {$ M = \text{tournament-selection}(P_{i+1})$}

		\STATE {$q_{i+1} = \text{crossover}(M) \cup \text{hierarchical-mutation}(M)  $ }
		\COMMENT {Check the FLOPS constraint at first (It takes $< 1ms$).}
		\IF {$FLOPS(q_{i+1}) > FLOPS_{max}$}
		\STATE {\textbf{continue}} 		 
		\ENDIF
		\STATE {Evaluate model $q_{i+1}$ with $S$ on $D$} 
		\COMMENT {Check the accuracy constraint  (It takes $\approx 60s$).}
		\IF {$Accuracy(q_{i+1}) > Acc_{min}$}
		\STATE {Add $q_{i+1}$ to $Q_{i+1}$}
		\ENDIF

		%			\FORALL { $q \in Q_{i+1}$}
		%				\STATE {Evaluate model $q$ with $S$ on $D$}
		%				\IF{$\text{Accuracy}(q) < C_{\text{Accuracy}}$}
		%					\STATE {$Q_{i+1} = Q_{i+1} \backslash \{ q \} $}
		%				\ENDIF
		%			\ENDFOR
		\ENDWHILE
		
		\ENDFOR
		\STATE {Select $K$ equispaced models near Pareto-front from  $P_{N}$}
	\end{algorithmic}
\end{algorithm}

\section{Experiments}

\subsection{Search Space}\label{sec:ss}
For later experiments, we add skip connections to commonly used search space to construct  $S_1$ and $S_2$. They are described as follows,

\textbf{Search Space $S_1$.} 
It is similar to ProxylessNAS \cite{cai2018proxylessnas}, where MobileNetV2 \cite{sandler2018mobilenetv2} is adopted as its backbone. In particular, $S_1$ is represented as a block-level supernet with $L=19$ layers of $N=7$ choices each. Its total size is $7^{19}$. The choices are,

\begin{itemize}
	\item MobileNetV2's inverted bottleneck blocks \cite{sandler2018mobilenetv2} of two expansion rates ($x$) in (3,6), three kernel sizes ($y$) in (3,5,7), labelled as MBE$x$K$y$\footnote{The order of numbering  $o=(x-3)+(y-3)/2$.}, 
	\item skip connection (the 6th choice\footnote{zero-based numbering}).
\end{itemize}

\textbf{Search Space $S_2$.} On top of $S_1$, we give each inverted bottleneck a squeeze-and-excitation \cite{hu2018squeeze} option (e.g., E$x$K$y$, E$x$K$y$\_SE), similar to MnasNet \cite{tan2018mnasnet}. Its total size thus becomes $13^{19}$.

We have to notice that skip connections are commonly used \cite{tan2018mnasnet,liu2018darts,bender2018understanding}, but meticulously neglected in recent single-path one-shot methods \cite{guo2019single,chu2019fairnas}. %This is however not a trivial decision. Later we show that including skip connections makes trouble for supernet training and also largely influences its ranking ability.

\subsection{NSGA-II Hyperparameters}
The hyperparameters for the weighted NSGA-II approach are given in Table~\ref{tab:pipelienhyper}.

\begin{table}[ht]	%\vskip 0.15in
	\begin{center}
		\begin{small}
				\begin{tabular}{|l|c|l|c|}
					\hline
					Item & value & Item & value \\
					\hline
					Population N & 70 & Mutation Ratio  & 0.8 \\
					$p_{rm}$ & 0.2 & $p_{re}$   & 0.65 \\
					$p_{pr}$ & 0.15 & $p_{M}$ & 0.7 \\
					$p_{K-M}$ & 0.3 & & \\
					\hline
				\end{tabular}
		\end{small}
	\end{center}
	\smallskip
	\caption{Hyperparameters for the weighted NSGA-II approach.}
	\label{tab:pipelienhyper}
	%\vskip -0.1in
\end{table}

\subsection{More Details about Scalable Supernet with ELS}
\begin{figure*}[ht]
	\centering
	\subfigure[First choice blocks' feature maps without ELS]{
		\includegraphics[scale=0.5]{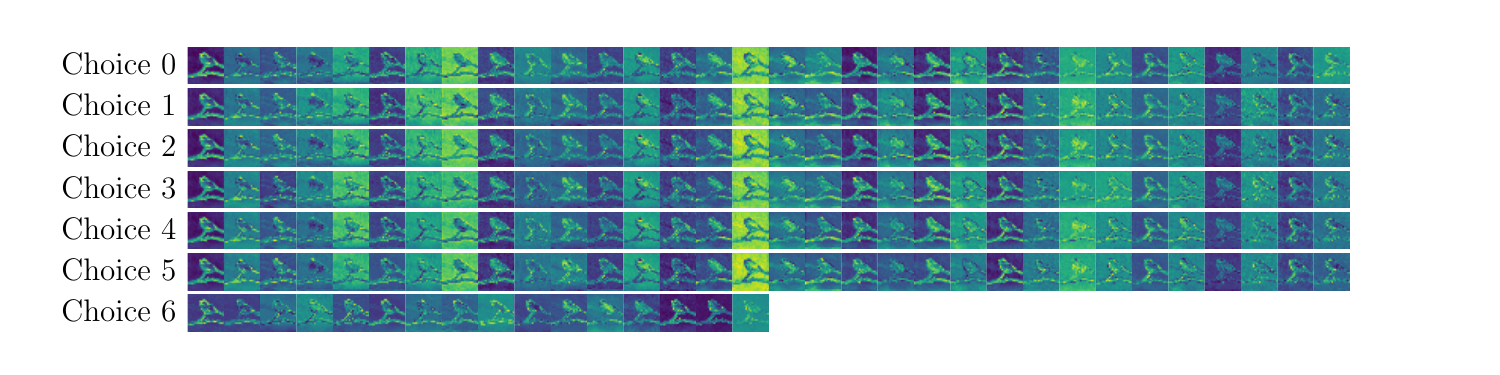}
	}
	\subfigure[First choice blocks' feature maps with ELS]{
		\includegraphics[scale=0.5]{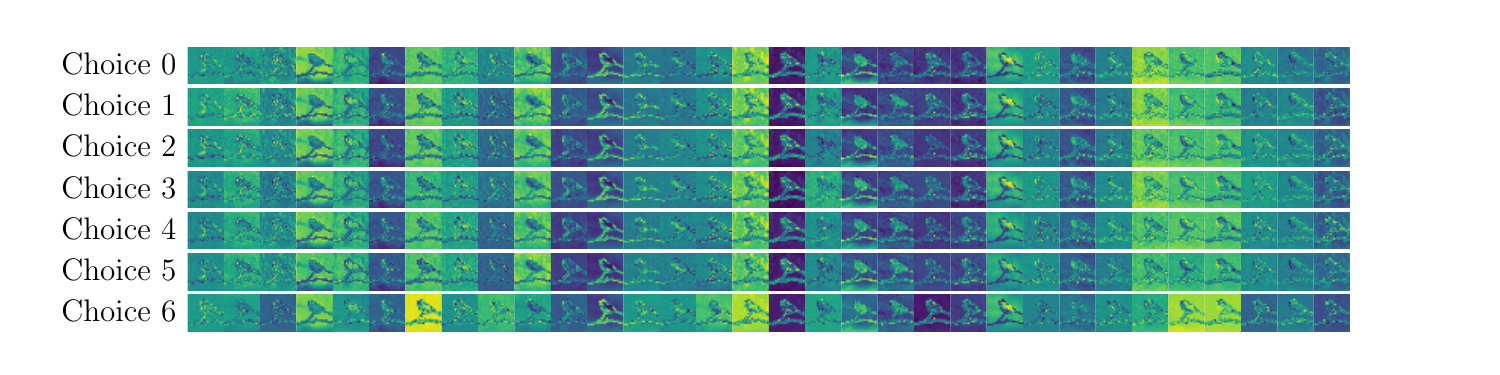}
	}
	\subfigure[High-level choice blocks' feature maps without ELS]{
		\includegraphics[scale=0.5]{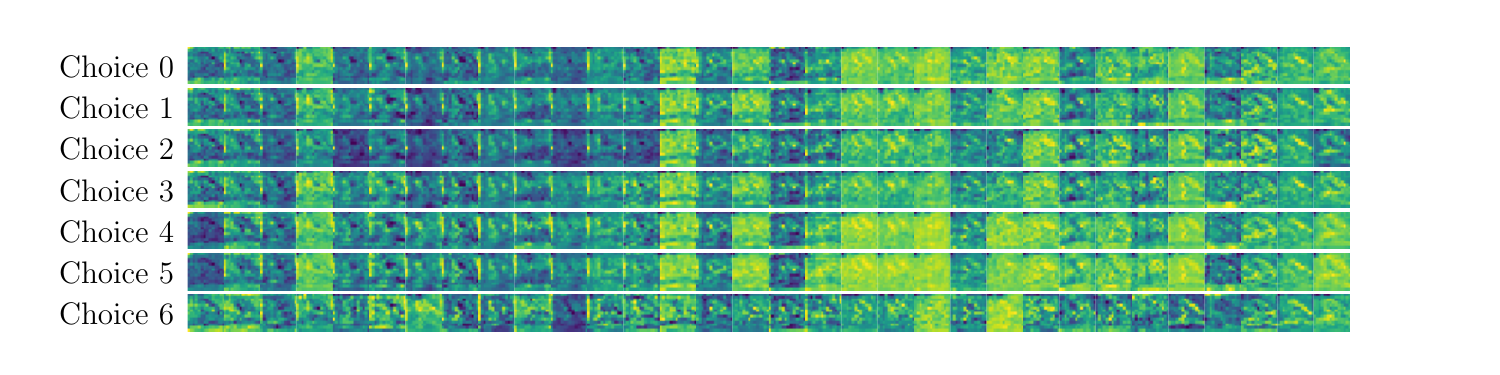}
	}
	\subfigure[High-level choice blocks' feature maps with ELS]{
		\includegraphics[scale=0.5]{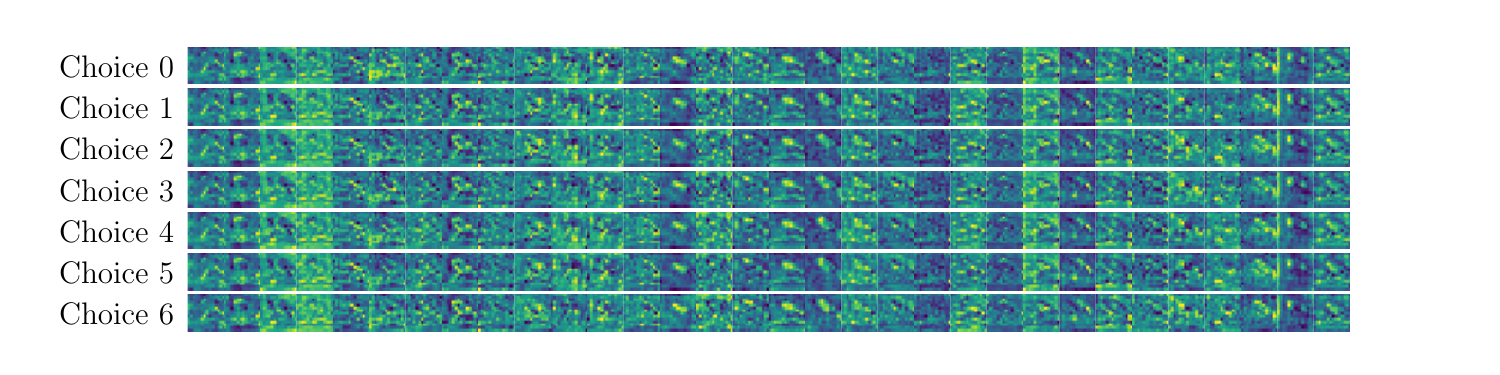}
	}
	\caption{Learned low-level and high-level features for the supernet with and without ELS.}
	\label{fig:bird-feature}
	\vskip -0.1in
\end{figure*}

Given an input of a chickadee\footnote{ImageNet ID: n01592084\_7680} image from ImageNet, we illustrate both high-level and low-level feature maps of the trained supernet with our proposed improvements in Figure~\ref{fig:bird-feature}. Pure skip connection easily interferes with the training process as it causes incongruence with other choice blocks. Note the channel size of feature map after Choice 6 in Figure~\ref{fig:bird-feature} (a) is half of others because the previous channel size is 16, while other choice blocks output 32 channels. This effect is largely attenuated by ELS. As it goes deeper, we still observe consistent high-level features. Specifically, when ELS is not enforced, high-level features of deeper channels easily get blurred out, while the supernet with ELS enabled continues to learn useful features in deeper channels. %We will discuss it formally in the next section.

\subsection{Search Space Evaluation}
NAS results can benefit from good search space. To prove the validity of the proposed method, we show our search space has a wide range and is not particularly designed. We pick two extreme cases, one with all identity blocks (only the stem and the tail remains),  the other with all K7E6s. They have the minimum and the maximum FLOPS respectively. We list their evaluation result in Table~\ref{tab:max-min-flops}.  The former has $24.1\%$ top-1 accuracy on ImageNet, and the latter $76.8\%$ at a cost of 557M FLOPs. Both are infeasible solutions as they violate either $acc_{min}$ or $madds_{max}$. It's thus a challenging task to deal with such search space for ordinary search techniques.

\begin{table*}
	\begin{center}
		\begin{small}
			\begin{tabular}{|l*{5}{|c}|} 			
				\hline
				Models & FLOPS (M)  & $>madds_{max}$   & Top-1  (\%) & Top-5  (\%) & $<acc_{min}$  \\
%				&  &   & (\%) & (\%) \\
				\hline
				All Identity &23  & No  &  24.1   & 45.0 & Yes  \\ % latest 6.9 % (Google) 
				All K7E6 & 557 & Yes  &76.8 & 93.3 & No \\ 
				\hline
			\end{tabular}
		\end{small}
	\end{center}
	\smallskip
	\caption{Full train results of models with minimal and  maximal FLOPS. } 
	\label{tab:max-min-flops}
\end{table*}

\subsection{Analysis of SCARLET Models}\label{sec:analysis-scarlet}

SCARLET-A makes full use of large kernels (five $5\times5$ and seven 7$\times$7 kernels)  to enlarge receptive field. Besides it activates many squeezing and excitation (12 out of 19) blocks to improve its classification performance. At the early stage, it appreciates either large kernels and small expansion ratios or small kernels and large expansion ratios to balance the trade-off between accuracy and FLOPs.

SCARLET-B chooses two identity operations. Compared with A, it shortens network depth at the last stages. Besides, it utilizes squeezing and excitation block extensively (14 out of 17). It places a large expansion block with large kernels at the tail stage.

SCARLET-C uses three identity operations and utilizes small expansion ratio extensively to cut down the FLOPs, large expansion ratio at the tail stage whose resolution is $7\times7$. It prefers large kernels before the downsampling layers. Besides, it makes an extensive use of squeeze and excitation to boost accuracy.

\end{document}